\newif\ifnarxiv

\narxivfalse

\ifnarxiv
\documentclass[9pt,twocolumn,twoside,lineno]{pnas-new}
\else
\documentclass[10pt]{article}
\usepackage[margin=1in]{geometry}
\usepackage[pdftex]{graphicx}
\usepackage{authblk}
\fi

\usepackage{amsmath, amssymb, bm}
\usepackage{color}

\ifnarxiv
\templatetype{pnasresearcharticle} 
\else
\begin{document}
\fi

\title{Model-Size Reduction for Reservoir Computing by Concatenating Internal States Through Time}


\ifnarxiv
\author[a,b,1]{Yusuke Sakemi}
\else
\author[a,b]{Yusuke Sakemi
\thanks{sakemi@iis.u-tokyo.ac.jp}
}
\fi
\author[a,c]{Kai Morino} 
\author[a,d]{Timoth\'{e}e Leleu}
\author[a,d]{Kazuyuki Aihara}

\affil[a]{Institute of Industrial Science, The University of Tokyo, 4-6-1 Komaba Meguro-ku, Tokyo 153-8505, Japan}
\affil[b]{NEC Corporation, 1753 Shimonumabe Nakahara-ku, Kanagawa 211-8666, Japan}
\affil[c]{Interdisciplinary Graduate School of Engineering Sciences, Kyushu University, 6-1 KasugaKouen, Kasuga-shi, Fukuoka 816-8580, Japan}
\affil[d]{International Research Center for Neurointelligence (WPI-IRCN), The University of Tokyo Institutes for Advanced Study, The University of Tokyo, Tokyo 113-0033, Japan}

\ifnarxiv
\leadauthor{Sakemi} 

\significancestatement{Reservoir computing can process time-series data efficiently. 
To employ reservoir computing in edge computing, it is highly desirable to reduce the amount of computational resources required.
In this study, we propose methods that can reduce the size of the reservoir up to one tenth without performance impairment. 
Because the proposed methods can be applied to any reservoir-computing model, the methods can be implemented in hardware such as FPGAs and photonic systems to enhance computational speed and energy efficiency.
}

\authorcontributions{All the authors designed the research. Y.S performed all simulations, and all the authors confirmed the theory. Further, they all wrote the paper.}
\authordeclaration{The authors have no conflict of interest to declare.}
\correspondingauthor{\textsuperscript{1} To whom correspondence should be addressed. E-mail: sakemi@iis.u-tokyo.ac.jp}


\fi

\ifnarxiv
\else
\maketitle
\fi

\begin{abstract}
Reservoir computing (RC) is a machine learning algorithm that can learn complex time series from data very rapidly 
based on the use of high-dimensional dynamical systems, such as random networks of neurons, called ``reservoirs.''
To implement RC in edge computing, it is highly important to reduce the amount of computational resources that RC requires.
In this study, we propose methods that reduce the size of the reservoir  
by inputting the past or drifting states of the reservoir to the output layer at the current time step. 
These proposed methods are analyzed based on information processing capacity, which is a performance measure of RC proposed by Dambre {\it et al.} (2012). 
In addition, we evaluate the effectiveness of the proposed methods on time-series prediction tasks: the generalized H\'{e}non-map and NARMA.
On these tasks, we found that the proposed methods were able to reduce the size of the reservoir up to one tenth without a substantial increase in regression error. 
Because the applications of the proposed methods are not limited to a specific network structure of the reservoir, 
the proposed methods could further improve the energy efficiency of RC-based systems, such as FPGAs and photonic systems.


\end{abstract}

\ifnarxiv
\dates{This manuscript was compiled on \today}
\doi{\url{www.pnas.org/cgi/doi/10.1073/pnas.XXXXXXXXXX}}

\begin{document}

\maketitle
\thispagestyle{firststyle}
\ifthenelse{\boolean{shortarticle}}{\ifthenelse{\boolean{singlecolumn}}{\abscontentformatted}{\abscontent}}{}
\fi

\ifnarxiv
\dropcap{E}fficiently
\else
Efficiently 
\fi
processing time-series data is important for various tasks, such as time-series forecasting, anomaly detection, natural language processing, and system control. 
Recently, machine-learning approaches for these tasks have attracted much attention of researchers and engineers because they not only require little domain knowledge but also often perform better than traditional approaches.
In particular, machine-learning models that employ recurrent neural networks, such as long short-term memory, have achieved great success in natural language processing and speech recognition \cite{Greff2017lstm}, and
their fields of applications continue to expand. 
However, the standard learning algorithms for recurrent neural networks, which include backpropagation through time \cite{Werbos1990backpropagation} and its variants \cite{Lillicrap2019backpropagation}, require large computational resources.
These computational burdens often hinder real-world applications, 
especially when computing is performed near end users. 
Such computing is called ``edge computing,'' and characterized by limited computational power and limited battery capacity. 

\begin{figure}
	\begin{center}
		\includegraphics[clip,width=7cm]{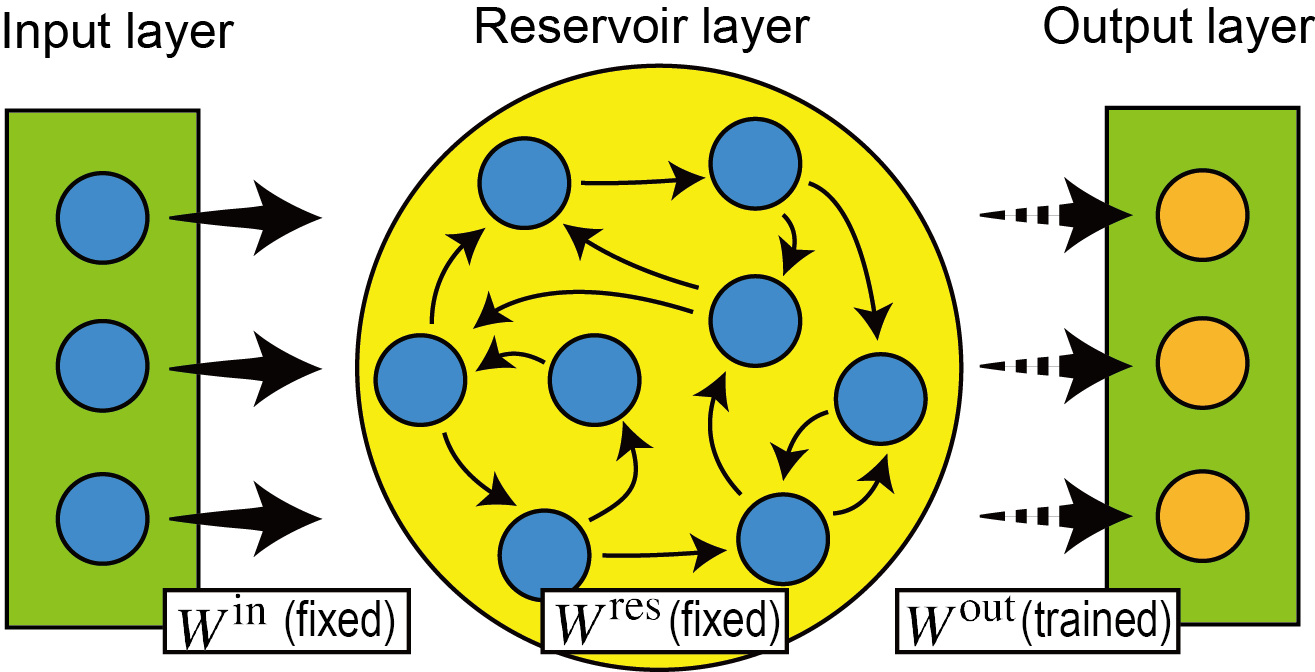}
		\caption{Typical RC architecture. The reservoir layer consists of randomly connected neurons. 
		The connections between the input and reservoir layer $W^\text{in}$ and connections within the reservoir layer $W^\text{res}$ are fixed (solid arrows), whereas the output weights $W^\text{out}$ are trained (dashed arrows).}
		\label{fig:RC_model}
	\end{center}
\end{figure}

Reservoir computing (RC) is a machine-learning algorithm that aims to reduce the computational resources required for predicting time series without reducing accuracy. 
As shown in Fig. \ref{fig:RC_model}, a typical RC consists of three parts: an input layer, 
a ``{\it reservoir}'' layer where neurons are randomly connected, 
and an output layer \cite{Jaeger2001echo,Maass2002realtime}.
Because only the weights between the reservoir layer and the output layer are trained while the other weights remain fixed, the learning process of RC is much faster than that of backpropagation through time \cite{Lukosevicius2009reservoir, Lukosevicius2012practical, Scardapane2017randomness}. 
Therefore, RC is expected to be a lightweight machine-learning algorithm that enables machine learning in edge computing \cite{Soures2017reservoir}.

The RC training is fast and accurate.
In addition, RC has shown high performance on various time-series forecasting tasks; examples include 
chaos \cite{Jaeger2004harnessing,Pathak2018model,Pathak2018hybrid}, weather \cite{McDermott2017ensemble}, 
wind-power generation \cite{Tian2019short}, and finance \cite{Lin2009short}.
Moreover, the range of applications of RC has extended into control engineering \cite{Tsai2010robust, Antonelo2015learning} and video processing \cite{Jalalvand2015real,Buteneers2013realtime,Panda2018learning}.

To develop the applications of RC in edge computing, its hardware implementation must be improved to enhance its computational speed and energy efficiency. 
For realizing such efficient hardware implementation, variants of RC models, some of which employ delay-feedback systems \cite{Appeltant2011information}, simple network topologies such as ring-topology and delay lines \cite{Ozturk2007analysis, Rodan2011minimum, Strauss2012design}, and billiard systems \cite{Katori2019reservoir}, have been proposed. 
Efficient hardware based on these variants have been implemented using field programmable integrated circuits (FPGAs) \cite{Alomar2015digital, Loomis2018fpga, Alomar2018efficient, Penkovsky2018efficient}.
Moreover, numerous types of implementation employing physical systems, such as photonics \cite{Brunner2018tutorial, Lima2017progress, Peng2018neuromorphic}, spintronics \cite{Torrejon2017neuromorphic}, mechanical oscillators \cite{Dion2018reservoir}, and analog integrated electronic circuits \cite{Bauer2019realtime, Yamaguchi2019chaotic}, have been demonstrated \cite{Tanaka2019recent}. 
Although these implementations have exhibited the superiority of RC in computational speed and energy efficiency, the maximum size of the reservoir is limited by the physical size of the hardware.

In this study, we propose three methods that reduce the size of the reservoir without any performance impairment.
The three methods share the concept that the number of the effective dimension of the reservoir is increased by allowing additional connections from the reservoir layer at multiple time steps to the output layer at the current time step. 
We analyze the mechanism of the proposed methods based on the {\it information processing capacity} (IPC) proposed by Dambre {\it et al.} \cite{Dambre2012information}.
We also demonstrate how the proposed methods reduce the size of the reservoir in the generalized H\'{e}non-map and NARMA tasks.

\section*{RC framework} \label{ss:methods}

In this section, we introduce the standard definition of RC and our proposed methods.

\subsection*{Definition of RC}

In the mathematical representation of RC, four vector variables are defined as follows: 
$\bm{u}(t) \in \mathbb{R}^{N^\text{in}}$ for the inputs, 
$\bm{x}(t) \in \mathbb{R}^{N^\text{res}}$ for the states of the reservoir, 
$\bm{y}(t) \in \mathbb{R}^{N^\text{out}}$ for the outputs, and 
and $\bm{y}^\text{tc}(t) \in \mathbb{R}^{N^\text{out}}$ for the teaching signals.
The constants $N^\text{in}, N^\text{res},$ and  $N^\text{out}$ are the dimensions of the inputs, states of the reservoir, and outputs, respectively. 
The updates of the reservoir states are given by 
\begin{flalign}
\bm{x}(t) &=  \tanh\left( W^\text{res}\bm{x}(t-1) + W^\text{in} \bm{u}(t)  \right),
\end{flalign}
where $W^\text{in}\in \mathbb{R}^{N^\text{res}\times N^\text{in}}$ is a weight matrix representing the connections from the neurons in the input layer to those in the reservoir layer. 
Its elements are independently drawn from uniform distribution $U(-\rho ^\text{in}, \rho ^\text{in})$, where $\rho ^\text{in}$ is a positive constant. 
Another weight matrix $W^\text{res}\in \mathbb{R}^{N^\text{res}\times N^\text{res}}$ represents the connections among the neurons in the reservoir layer. 
Its elements are initialized by drawing values from uniform distribution $U(-1, 1)$ and subsequently divided by a positive value  
to ensure that the spectral radius of $W^\text{res}$ is $\rho ^\text{res}$. 
Note that elements in matrices $W^\text{in}$ and  $W^\text{res}$ are fixed to the initialized values.  
The outputs are obtained by
\begin{flalign}
\bm{y}(t) =W^\text{out}\bm{x}(t),
\end{flalign}
where $W^\text{out}\in \mathbb{R}^{N^\text{out} \times N^\text{res}}$ is a weight matrix representing connections from the neurons in the reservoir layer to those in the output layer. 
The output weight matrix $W^\text{out}$ is trained in the offline learning process of RC by using the pseudoinverse (see Materials and Methods sections). 

\subsection*{Proposed methods}

\begin{figure*}
	\begin{center}
		\includegraphics[clip,width=14cm]{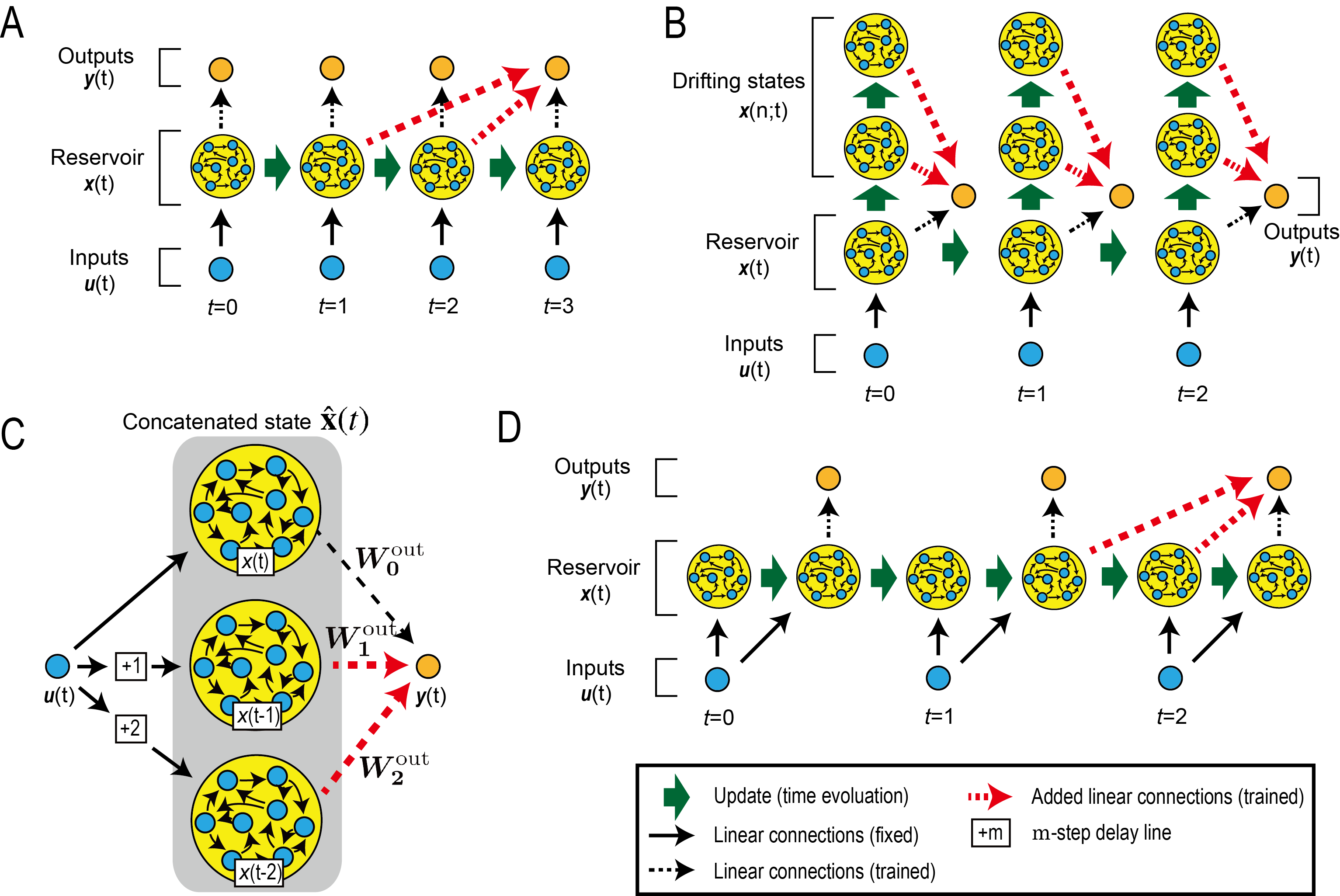}
		\caption{Schematic of the proposed methods. A. Delay-state concatenation  
		when the number of additional connections $P$ is two and the unit of delay $Q$ is one.  
		B. Drift-state concatenation when $P$ is two and $Q$ is one.
		C. Another view of delay-state concatenation. 
		The reservoir consists of three identical dynamical systems and delay lines. 
		The added dynamical systems have $+1$ delay lines and $+2$ delay lines, respectively. 
		D. Delay-state concatenation with one transient state 
		when $P$ is two and $Q$ is one.  
		}
		\label{fig:proposed}
	\end{center}
\end{figure*}

We propose three methods that modify the connections between the reservoir and output layers. 
We call these three methods (i) {\it delay-state concatenation},  (ii) {\it drift-state concatenation}, and 
(iii) {\it delay-state concatenation with transient states}. 
These methods share the idea that the number of the effective dimension of the reservoir is increased by allowing additional connections 
from the reservoir layer at multiple time steps to the output layer at the current time step. 
For the delay-state concatenation and delay-state concatenation with transient states, 
additional connections are formed from the past states of the reservoir layer to the current output layer, 
as illustrated in Figs. \ref{fig:proposed} A and D.
On the other hand, for the drift-state concatenation, 
additional connections are formed from newly introduced states of the reservoir, called drifting states,  
to the current output layer, as illustrated in  Fig. \ref{fig:proposed} B.
The drifting states are obtained by updating the current states of the reservoir layer without input signals.  
In the following section, we mathematically formulate these three proposed methods. 

First, we formulate the delay-state concatenation with concatenated states of the reservoir given by 
\begin{flalign}
\bm{\hat{x}}(t) &:= \left( \begin{array}{c}
\bm{x}(t) \\
\bm{x}(t-Q)\\
\vdots \\
\bm{x}(t-PQ)
\end{array}
\right).
\end{flalign}
Note that $\bm{x}(t)$ is a column vector and the number of neurons in the reservoir does not change. 
A positive integer $Q$ represents the unit of delays.
Another positive integer $P$ represents the number of past states that are concatenated to the current states.
The outputs are obtained with the concatenated states $\bm{\hat{x}}(t)$ and the corresponding output-weight matrix $\hat{W}^\text{out}$ as follows: 
\begin{flalign}
\hat{W}^\text{out} &:= \left( W^\text{out}_0~W^\text{out}_1~\cdots~W^\text{out}_P \right), \\
\bm{y}(t)&= \sum _{i=0}^{P} W_i ^\text{out} \bm{x}(t - iQ) \nonumber \\
&=\hat{W}^\text{out}\bm{\hat{x}}(t).
\end{flalign}
Here, $\bm{\hat{x}}(t)$ and $\hat{W}^\text{out}$ are defined in $\mathbb{R}^{(P+1) N^\text{res}}$ and $\mathbb{R}^{N^\text{out} \times (P+1) N^\text{res}}$, respectively.
Figure \ref{fig:proposed} A shows a schematic of this method illustrating the prediction of $\bm{y}(3)$ when $Q=1\text{ and }P=2$. 
One can see that there are additional connections from the past states of the reservoirs  $\bm{x}(1)$ and $\bm{x}(2)$ to output $\bm{y}(3)$, as indicated by red dashed arrows.
From a different point of view, this model can be illustrated using the concatenated states of the reservoir $\hat{\bm{x}}(t)$, as in Fig. \ref{fig:proposed} C, where 
the reservoir consists of three identical smaller reservoirs, each with a different time delay of $0$, $1$, and $2$ from the inputs.
Evidently, the effective dimension of the concatenated reservoir is three times larger than that of the original reservoir. 
Because the learning performance is enhanced by using a larger reservoir \cite{Lukosevicius2012practical}, 
the proposed method should be able to increase the computing capability without needing to add neurons in the reservoir.

Second, we formulate drift-state concatenation, as illustrated in Fig. \ref{fig:proposed} B 
by introducing the drifting states of the reservoir given by 
\begin{flalign}
\bm{x}^\text{drift}(t';t) &=\begin{cases}
 \tanh\left( W^\text{drift}\bm{x}(t) \right), ~(\text{if }t'=1), \\
 \tanh\left( W^\text{drift}\bm{x}^\text{drift}(t'-1;t) \right), ~(\text{if }t'\ge2), \\
 \end{cases} 
\end{flalign}
where $\bm{x}^\text{drift}(t';t)$ represents the drifting states of the reservoir and 
$t'$ is the time step after the current time step $t$. 
Using the drifting states, we redefine the concatenated states of the reservoir and the corresponding output matrix as follows:
\begin{flalign}
\bm{\hat{x}}(t) &:= \left( \begin{array}{c}
 \bm{x}(t) \\
 \bm{x}^\text{drift}(1;t)\\
 \bm{x}^\text{drift}(2;t)\\
 \vdots \\
 \bm{x}^\text{drift}(P;t)
\end{array}
\right),  \\
\hat{W}^\text{out} &:= \left(W_0^\text{out } W_1^\text{out } \cdots W_P^\text{out} \right), \\
\bm{y}(t)&= W_0^\text{out} \bm{x}(t) + \sum _{i=1}^{P} W_i^\text{out} \bm{x}^\text{drift}(i;t) \nonumber \\
& =\hat{W}^\text{out}\bm{\hat{x}}(t). 
\end{flalign}
Here, $W^\text{drift}$ is a matrix representing the connections within the reservoir to obtain the drifting states,  
and its elements are drawn from uniform distribution $U(-1, 1)$ divided by a positive value  
to ensure that the spectral radius of $W^\text{drift}$ is $\rho ^\text{drift}$.

Third, delay-state concatenation with transient states introduces transient states to the delay-state concatenation, as illustrated in Fig. \ref{fig:proposed} D  
where the states of the reservoir update twice (so it has one transient state) 
during the inputs and the outputs update once (see Materials and Methods section). 

Although we have shown that the proposed methods can increase the effective dimension of the reservoir without adding neurons, 
one potential drawback of the methods is the cost of the memory required to store the past reservoir states. 
Against expectation, however, delay-state concatenation is very memory efficient, as explained below. 

To carry out RC with delay-state concatenation, the terms 
\begin{flalign}
W_0^\text{out} \bm{x}(t),~W_1^\text{out} \bm{x}(t), \dots, \text{ and } W_P^\text{out} \bm{x}(t) 
\end{flalign}
are computed and stored in memory at time step $t$.  
The dimensions of these vectors are all $N^\text{out}$.
The vector $W_i^\text{out} \bm{x}(t)$ must be stored until they are used for calculating the outputs at time steps $t+iQ$. 
Therefore, the total memory cost is obtained as follows: 
\begin{flalign}
&N^\text{out} + (Q+1)N^\text{out} + (2Q+1)N^\text{out} + \cdots + (PQ+1)N^\text{out} \nonumber \\
&= \frac{(P+1)(PQ+2)N^\text{out}}{2}. \label{eq:MemoryCost}
\end{flalign}
It should be noted that this method increases the effective dimension of the reservoir state by a factor of $(P+1)$, but the number of the neurons in the reservoir is not increased. 
If the number of neurons is increased by $(P+1)$ times, the memory cost to store the reservoir states becomes $(P+1)N^\text{res}$. 
For moderate numbers of $P$ and $Q$ (typically less than 5), 
the memory cost for the proposed method given by Eq. \ref{eq:MemoryCost} is much less than $(P+1)N^\text{res}$ because $N^\text{out} \ll N^\text{res}$ for typical RC applications.
The memory required to store the weights within the reservoir is also reduced from $\left(PN^\text{res}\right)^2 + PN^\text{out}$ to $\left(N^\text{res}\right)^2 + PN^\text{res}N^\text{out}$ in the same way given that the reservoir is fully connected.
Therefore, delay-state concatenation can increase the dimensions of the reservoir more efficiently than by simply increasing the number of neurons in the reservoir.

\section*{Quantitative analysis based on IPC} \label{ss:IPC}

Before benchmarking the proposed RC, 
we quantitatively analyze the learning capacity of the RC  
to elucidate how the proposed methods work.

The {\it memory capacity} (MC) is a performance measure commonly used in the RC research community \cite{Jaeger2001short}.
The MC represents how precisely the RC system can reproduce the past inputs. 
A number of studies have shown that the MC  
is theoretically bounded by the number of neurons in the reservoir, 
and the MC can reach this bound in some situations \cite{White2004short, Ganguli2008memory, Rodan2011minimum, Strauss2012design}. 
Boedecker {\it et al. }\cite{Boedecker2012information} evaluated the MC at the edge of chaos, which is a region in the model parameter space where RC is stable but near to unstable.  
Farkaš {\it et al. }\cite{Farka2016computational} have evaluated the MC for various model parameters. 
However, the MC does not evaluate how well RC processes information in a nonlinear way. 
Because many tasks in the real world targeted by RC are nonlinear problems, 
the MC is not a suitable measure for analyzing the proposed methods in this sense. 
Therefore, to elucidate the mechanism of the methods proposed in this study, 
we employed another criterion \cite{Dambre2012information} called the information processing capacity (IPC),  
which handles nonlinear tasks.

The IPC is a measure that integrates both memory and information processing performance. 
By employing an orthogonal basis set that spans the Hilbert space, 
one IPC can be obtained from one corresponding basis. 
The IPC can be interpreted as a quantity that represents not only how well the network can memorize past inputs but also how precisely the network can convert inputs into the target outputs in a nonlinear manner given the basis set.  
Dambre {\it et al. } \cite{Dambre2012information} showed that the total IPC $C^\text{total}$, which is a sum of all IPCs, 
is identical to the number of neurons in the reservoir, provided 
(\MakeUppercase{\romannumeral 1}) the inputs are independent and identically distributed ({\it i.i.d.}), 
(\MakeUppercase{\romannumeral 2}) the fading memory condition is satisfied, 
and (\MakeUppercase{\romannumeral 3}) all the neurons are linearly independent 
(see Theorem 7 and its proof in \cite{Dambre2012information}). 
By analyzing the IPCs, one can obtain a large amount of information about how RC processes input data.  
For example, the degree of nonlinearity of the information processing carried out in RC can be analyzed by calculating multi-order IPCs. 
The $k$th-order IPC $C^\text{$k$th}$ is defined as the sum of the IPCs corresponding to the subset of a basis with $k$th-order nonlinearity.  
Based on the IPC, informative results such as the memory--nonlinearity tradeoff have been obtained \cite{Dambre2012information}. 
Therefore, using the IPC, we can analyze how RC stores and processes information in the reservoir as well as how the proposed methods affect the way information is processed.

\begin{figure*}
	\begin{center}
		\includegraphics[clip,width=16cm]{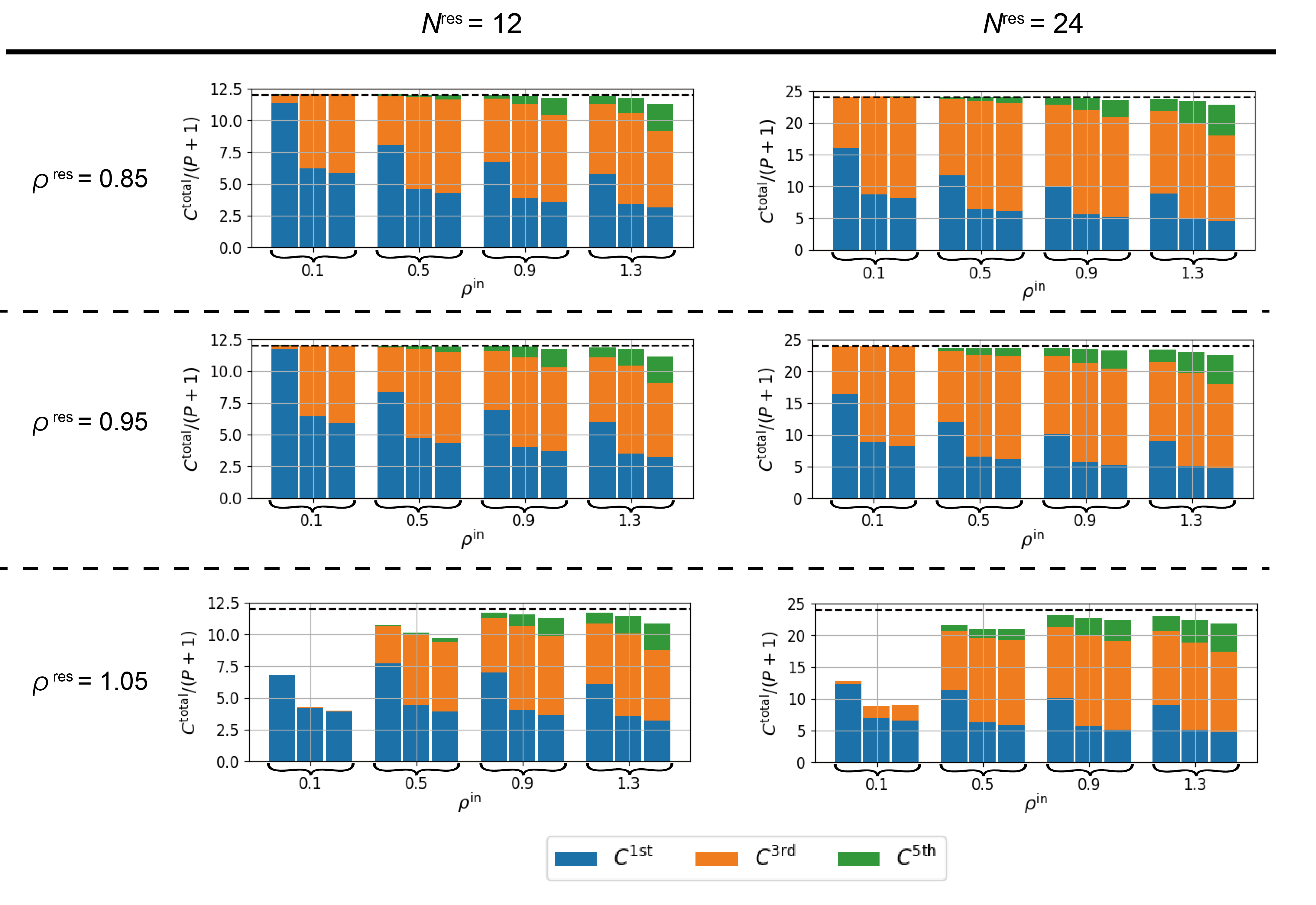}
		\caption{Information processing capacities (IPCs) of standard RC and RC with the proposed methods for various values of input weight strength $\rho ^\text{in}$ and spectral radius $\rho ^\text{res}$.
		The left and right panels show the results for $N^\text{res}=12$ and $N^\text{res}=24$ , respectively.
		The dashed horizontal line within each subgraph represents the value of $N^\text{res}$.
		For each value of $\rho ^\text{in}$, the left column presents the standard RC with $P=0$, the center column presents delay-state concatenation with $P=Q=1$, and the right column presents drift-state concatenation with $P=Q=1$.}
		\label{fig:Dambre_bar}
	\end{center}
\end{figure*}

We calculated the IPCs for the standard RC and for RC with the proposed methods (see Materials and Methods section). 
Figure \ref{fig:Dambre_bar} shows the IPCs when $N^\text{res}=12$ and $N^\text{res}=24$ for various values of $\rho^\text{in}$ and $\rho^\text{res}$.
Note that only odd-order IPCs were observed because of the symmetry. 
In each setting, we calculated the IPCs for the standard RC (left columns), those for the RC with delay-state concatenation with $P=Q=1$ (center columns), and those for the RC with drift-state concatenation with $P=Q=1$ (right columns).
One can find that the value of $C^\text{total}/(P+1)$ almost reaches the number of neurons $N^\text{res}$ in reservoir except when $\rho^\text{res}=1.05$. 
This result indicates that the proposed methods actually increase the total IPC by $(P+1)$ times. 
The observed lower values of the total IPCs for the case of $\rho^\text{res}=1.05$ can be attributed to the failure-of-fading-memory condition. 
In the RC research community, it is well known that the dynamics of RC is more likely to be chaotic when $\rho^\text{res}$ increases (typically occurring when $\rho^\text{res}$ is larger than $1$) \cite{Yildiz2012revisiting}, and this corresponds to the failure of fading memory. 
For all cases, as $\rho ^\text{in}$ increases, the third-order IPC $C^{3\text{rd}}$ and the fifth-order IPC $C^{5\text{th}}$ tend to increase, 
which reflects the increase in nonlinearity in the reservoir \cite{Dambre2012information} given the selected basis set.
Note that as $\rho ^\text{in}$ increases, the total IPC $C_\text{total}$ tends to decrease, 
which is attributed to increases in the importance of higher-order IPCs (e.g., seventh-order and ninth-order IPCs).

\begin{figure*}
	\begin{center}
		\includegraphics[clip,width=18cm]{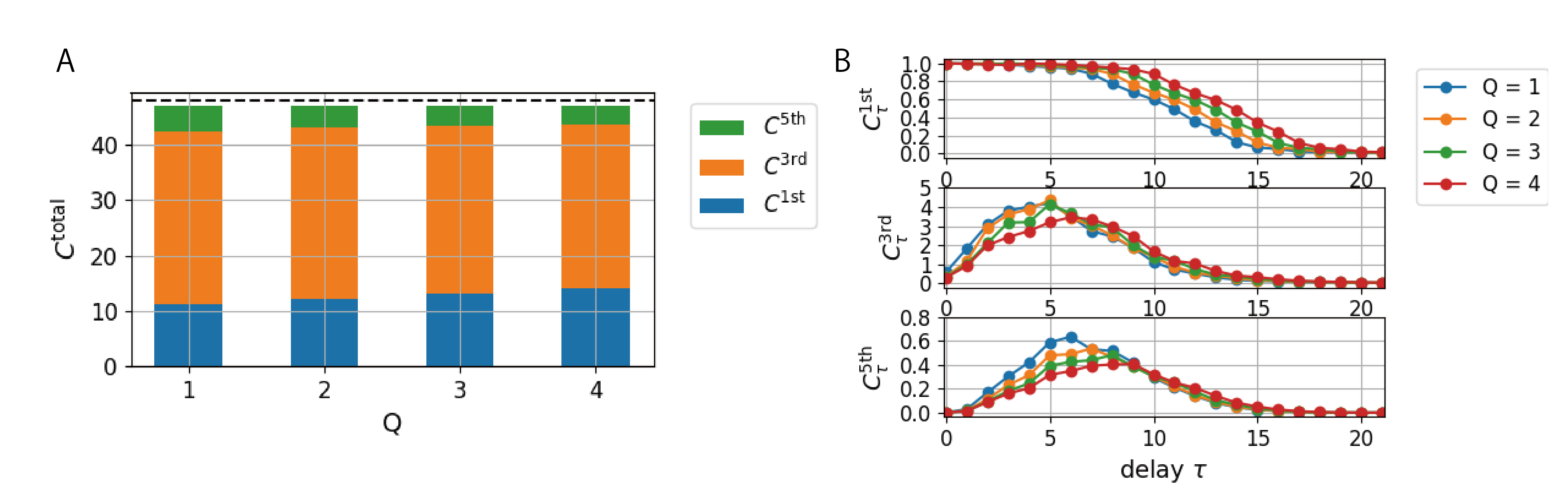}
		\caption{(A) IPCs of the RC with delay-state concatenation for various values of $Q$.
(B) Delay structures of the IPCs for various values of $Q$.
		From top to bottom, the first-, third-, and fifth-order IPCs are shown.
	The parameters are set as follows: $N^\text{res}=24$, $\rho ^\text{in} = 0.9$, $\rho ^\text{res} = 0.95$, and $P=1$.}
		\label{fig:Dambre_Q}
	\end{center}
\end{figure*}

In Fig. \ref{fig:Dambre_Q} (A), we show the IPCs for delay-state concatenation with $P=1$ for several values of the unit of delay $Q$.
As $Q$ increases, the first-order IPC increases as well. 
This result may be trivial because RC with large $Q$ can access the past states of the reservoir, rendering the reproduction of the past inputs easy. 
To investigate the effects of the unit of delay $Q$ on IPCs, 
we decomposed the $k$th-order IPC $C^{k\text{th}}$ into components in terms of their delay such as $C_0^{k\text{th}},~C_1^{k\text{th}},~\text{and }C_2^{k\text{th}}$, which correspond to a different subset of the basis.   
Figure \ref{fig:Dambre_Q} (b) shows the distributions of the delay components for four values of $Q$ under the same experimental conditions. 
As the values of $Q$ increase, the distribution tends to shift to the right (larger delays) for each order of IPC. 
This fact indicates that, as demonstrated in the subsequent section,  one can tune RC models by adjusting the value of $Q$ according to the delay structure of the target tasks. 

\begin{figure*}
	\begin{center}
		\includegraphics[clip,width=17cm]{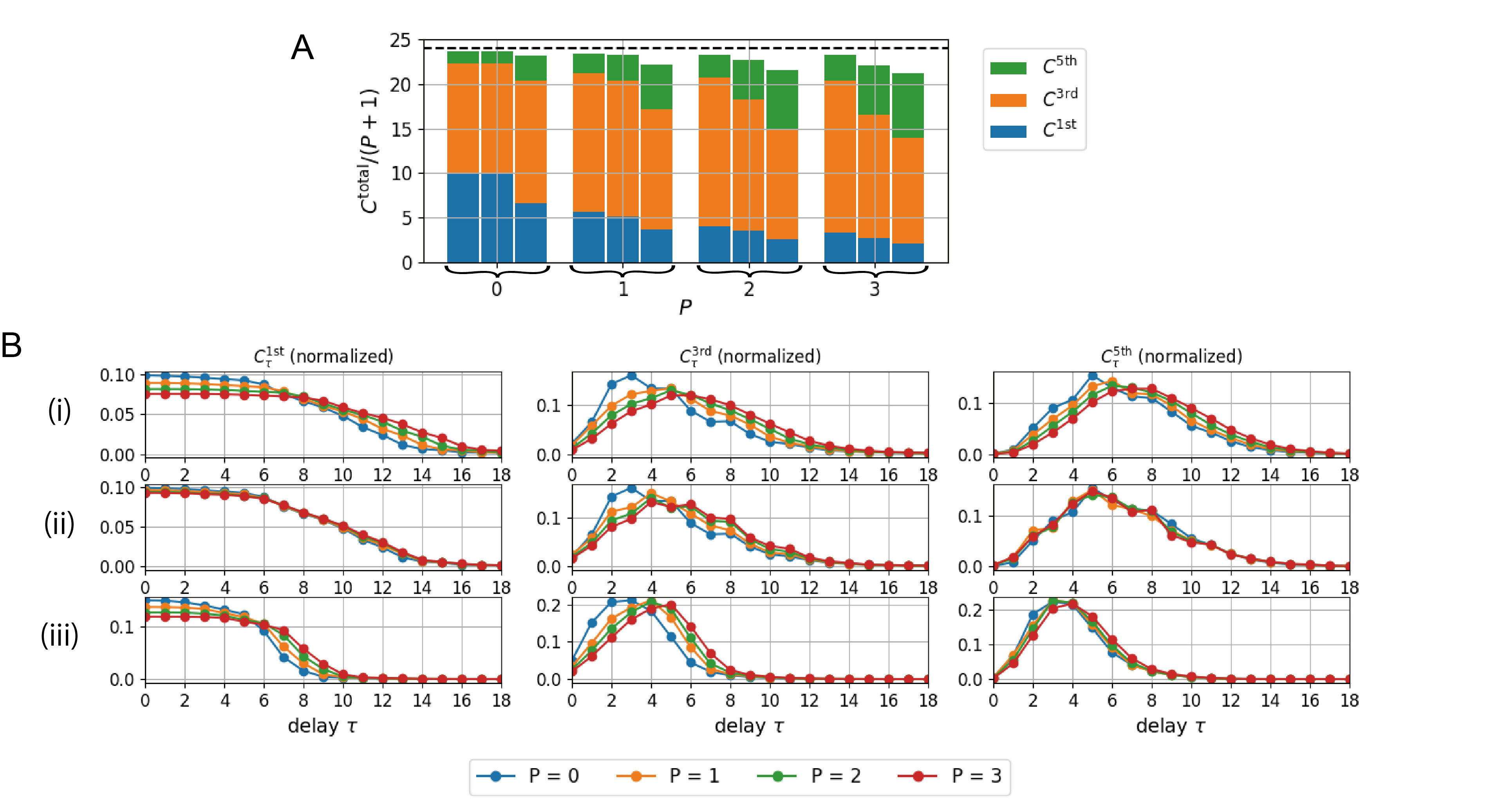}
		\caption{(A) IPCs with the proposed methods for various values of $P$.
		Left columns: IPCs for delay-state concatenation, 
		center columns: IPCs for drift-state concatenation, 
		and right columns: IPCs for delay-state concatenation with one transient state.
		(B) Comparison of the delay structures of the IPCs for the three proposed methods.
		Top panels (i):  delay structures for delay-state concatenation,
	middle panels (ii): delay structures for drift-state concatenation,
 	and bottom panels (iii): delay structures for delay-state concatenation with one transient state.
		To highlight the difference in the distribution of the delay structures, the IPCs for each order are normalized.
		From left to right, the first-, third-, and fifth-order IPCs are shown.	
		The parameters are set as follows: $N^\text{res}=24$, $\rho ^\text{in} = 0.9$, $\rho ^\text{res} = 0.95$, and $Q=1$.}
		\label{fig:Dambre_P}
	\end{center}
\end{figure*}

Next, for various values of $P$, we show the IPCs in Fig.  \ref{fig:Dambre_P} (A) for delay-state concatenation, drift-state concatenation, and delay-state concatenation with one transient state.
For all three proposed methods, the contributions of higher-order IPCs tend to be dominant in the total IPCs as $P$ increases. 
In Fig. \ref{fig:Dambre_P} (B), we show the delay structures of the IPCs. 
Note that to clarify how the distributions of the delay structure change as the value of $P$ changes, 
we used the normalized IPC $C^{n\text{th}}_\tau / (P+1) \sum _\tau C^{n\text{th}}_\tau$. 
The top panels in Fig. \ref{fig:Dambre_P} (B) show that, as $P$ increases, the distribution of the delay structure of the IPCs for delay-state concatenation tends to shift to the right (larger delays).   
Conversely, in the middle panels, the IPCs for drift-state concatenation do not change significantly. 
These results may be explained as follows: 
the increase in $P$ for delay-state concatenation increases the memory of past inputs because of the additional connections from the past states of the reservoir,  
whereas the increase in $P$ for drift-state concatenation does not increase the memory of the past inputs because drifting states are obtained from the current states of the reservoir.  
The bottom panels of this figure show that the distribution of IPCs for delay-state concatenation with one transient state tends to shift to larger delays as $P$ increases.  
However, the delays in this distribution are smaller than the delays in the distribution obtained using delay-state concatenation. 
This difference may stem from the fact that the information of past states is more likely to be thrown away in delay-state concatenation with one transient state 
because the RC model in this case carries out nonlinear transformation twice for each input (see Fig. \ref{fig:proposed} D).

Here, we present a short summary of the above experiments.
We have numerically shown that the total IPCs divided by $P+1$ are almost independent of the values of $Q$ and $P$, 
which is consistent with the theory in Ref. \cite{Dambre2012information}. 
Furthermore, we have found that the importance among IPC components and the delay structure of IPCs 
can be modified by selecting the values of $Q$ and $P$.
These findings indicate that the learning performance on real-world tasks may be enhanced by selecting appropriate values of $Q$ and $P$ 
adjusted to a target task with a specific temporal structure. 

\section*{Effectiveness on Complex Data} \label{ss:applications}

Although we have shown that the proposed methods can increase the IPCs efficiently, 
the conditions assumed above are not always guaranteed in real-world applications; for example, inputs may not be drawn from {\it i.i.d.} data, and neurons in the reservoir may not be linearly independent. 
Therefore, the IPCs are just a guide that help us understand the mechanisms of the proposed methods. 
In this section, to evaluate the effectiveness of the proposed methods on complex data, we applied them to two prediction tasks: generalized H\'{e}non-map tasks and NARMA tasks (see Eq. \ref{eq:Henon} and Eq. \ref{eq:NARMA} in Materials and Methods section).
In the following experiments, the dimensions of the reservoir are approximately given by an integer $N^*$, and the actual number of neurons in the reservoir is given by $N^\text{res} = \lfloor N^* / (P+1) \rfloor$, where $\lfloor m \rfloor := \text{max} \{q\in \mathbb{Z}|q\le m\}$.
We optimized the model parameters, $\rho ^\text{in},~\rho ^\text{res}$, and $\rho ^\text{drift}$ with Bayesian optimization \cite{Snoek2012practival,Frazier2018tutorial}.

\begin{figure}
	\begin{center}
		\includegraphics[clip,width=8cm]{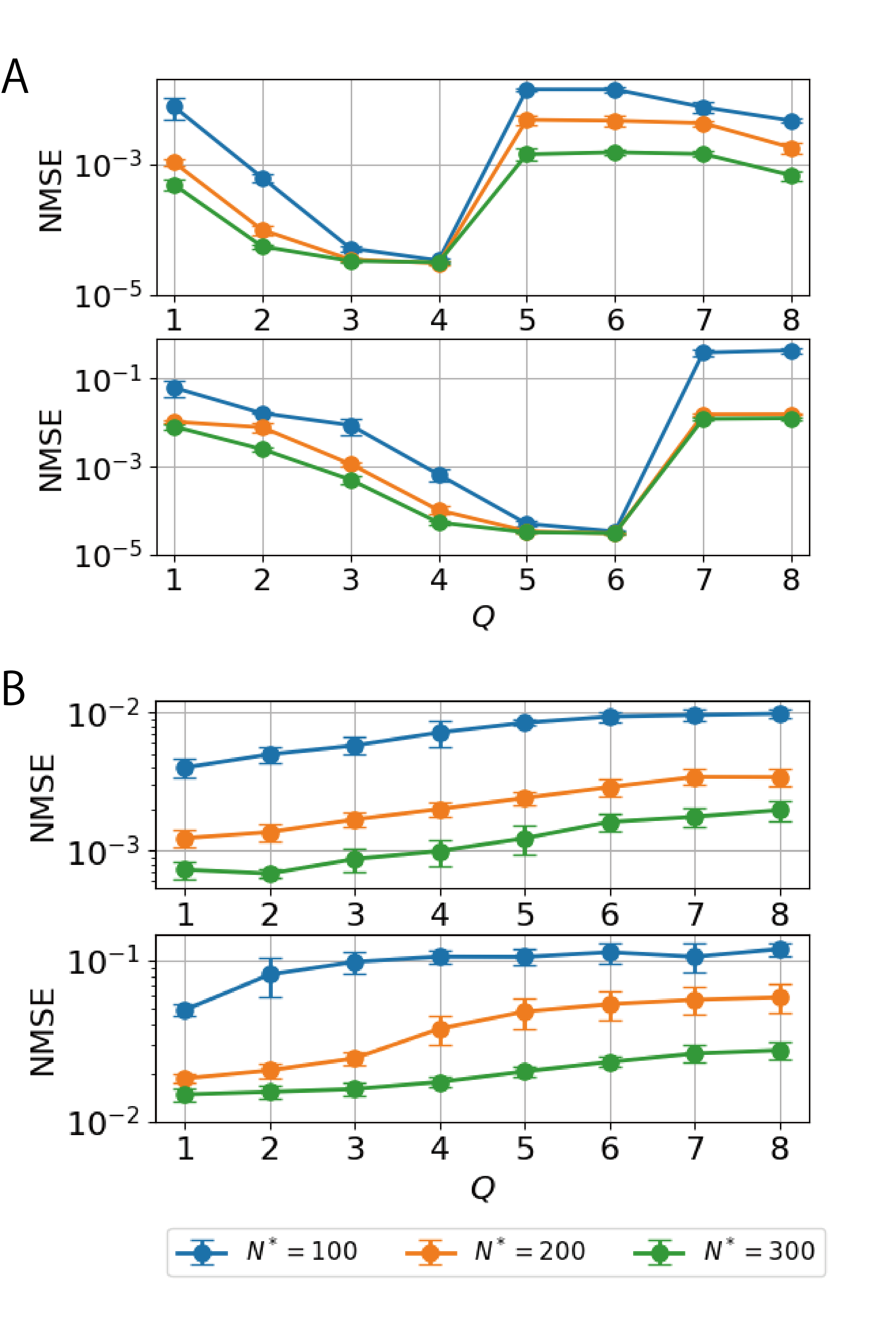}
		\caption{(A) NMSE for a sixth-order H\'{e}non-map task (the upper panel) and for an eighth-order H\'{e}non-map task (the lower panel) for various values of $Q$ with $P=1$. 
		(B) NMSE for NARMA5 (the upper panel) task and for NARMA10 (the lower panel) task for various values of $Q$ with $P=1$.  Error bars show the standard deviation of the results of 10 trials. }
		\label{fig:results_Q}
	\end{center}
\end{figure}

We first investigated the effects of the value of $Q$ in the delay-state concatenation method.
Figure \ref{fig:results_Q} (a) shows the normalized mean-squared errors (NMSEs) for the sixth-order and eighth-order H\'{e}non-map tasks 
for various values of $Q$ with $P=1$.
As the value of $Q$ increases, 
the NMSE first decreases, but abruptly increases when $Q$ is larger than 4 in the sixth-order H\'{e}non-map and larger than 6 in the eighth-order H\'{e}non-map.
Considering the fact that to predict the output, the $m$th-order H\'{e}non-map has two informative inputs at the $m$th and $(m-1)$th previous steps,  
these increases in performance are reasonable because a RC system with the appropriate values of $Q$ can possess the information needed from past inputs. 
We also note that similar results were obtained recently in \cite{Marquez2019Takens}.
In contrast,  
the NMSE monotonically increases as the value of $Q$ increases for NARMA5 and NARMA10, as shown in Fig. \ref{fig:results_Q} (b).
These results imply that simply adjusting the value of $Q$ is not effective 
for tasks such as NARMA5 and NARMA10 with complicated temporal structures.

\begin{figure}
	\begin{center}
		\includegraphics[clip,width=8cm]{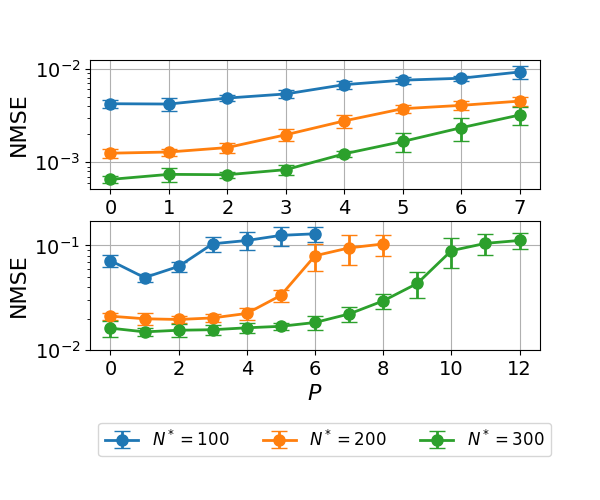}
		\caption{NMSEs for the NARMA5 (the top panel) and NARMA10 (the bottom panel) tasks with delay-state concatenation as functions of $P$ for various values of $N^*$. Error bars show the standard deviation of the results of 10 trials. }
		\label{fig:NARMA_P}
	\end{center}
\end{figure}

We next investigated the effects of the value of $P$ for the NARMA tasks in which adjusting the value of $Q$ was not effective.
Figure \ref{fig:NARMA_P} shows the NMSE against the values of $P$ for the NARMA5 task (the top panel) and NARMA10 task (the bottom panel) 
for three settings of $N^*$.
Note that $N^*$ is fixed as $P$ is varied
to ensure that the size of reservoir $N^\text{res}$ reduces to approximately $1/(P+1)$ times as $P$ varies. 
We found that the NMSEs were almost constant up to specific values of $P$. 
For example, for the case of the NARMA10 task with $N^*=300$, the NMSE was almost constant  up to $P=5$, 
indicating that the number of neurons in the reservoir can be reduced from 300 to 50 without impairing performance.

\begin{figure}
	\begin{center}
		\includegraphics[clip,width=8cm]{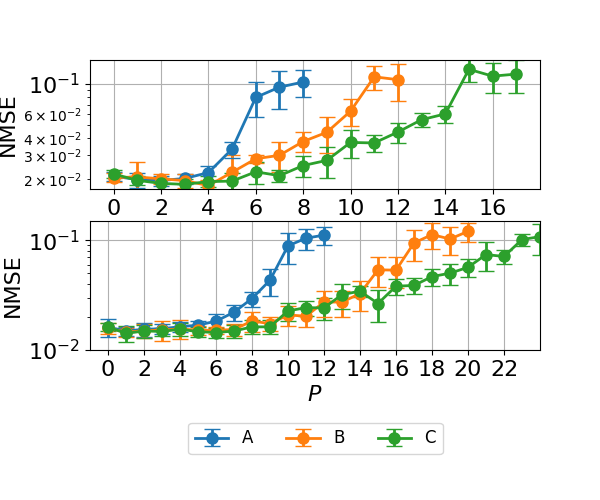}
		\caption{Results of regression performance on the NARMA10 task for various values of $P$ 
		with $N^* = 200$ (the top panel) and $N^* = 300$ (the bottom panel) for (A) delay-state concatenation, (B) drift-state concatenation, and (C) delay-state concatenation with one transient state. 
		Error bars show the standard deviation of the results of 10 trials. 
		}
		\label{fig:NARMA_methods}
	\end{center}
\end{figure}

Finally, we compared the proposed methods, delay-state concatenation with and without one transient state, and drift-state concatenation on the NARMA10 task.
Figure \ref{fig:NARMA_methods} shows the NMSE  as functions of $P$ values with $N^*=200$ and $300$ for the proposed methods.
We found that the NMSEs for delay-state concatenation with one transient state and drift-state concatenation were lower 
than those for delay-state concatenation when $P$ was larger than 7.
For the NARMA10 task, inputs more than 10 steps in the past are not very informative for predicting one step forward. 
The lack of information in the past steps may explain the increase in NMSE for delay-state concatenation when $P$ is larger than 7. 
The observed lower NMSEs for delay-state concatenation with one transient state and drift-state concatenation are also reasonable because 
(\MakeUppercase{\romannumeral 1}) the former method uses the past states of the reservoir, which contain more recent input information for the same value of $P$ (see Fig. \ref{fig:proposed} D), and  
(\MakeUppercase{\romannumeral 2}) the latter method uses the states of the reservoir, which contain current input information.

\section*{Discussion} \label{ss:discussion}

In this study, we proposed three methods to reduce the size of an RC reservoir without impairing performance.

To elucidate the mechanism of the proposed methods,  
we analyzed the IPC.
We found that the value of the total IPCs almost reaches $N^\text{res} (P+1)$ using the proposed methods, whereas the importance of their components (the first-, third-, and fifth-order IPCs) changes drastically.
We also found that the delay structures of the IPCs depend on the value of $Q$ and $P$. 
To investigate the applicability of the proposed methods on complex data, 
we presented the experimental results on generalized H\'{e}non-map and NARMA tasks.  
We found that when the target task has a relatively simple temporal structure, as demonstrated with the H\'{e}non-map tasks,   
selecting an appropriate value of $Q$ enhances the performance substantially.  
In contrast, when the target task contains complex temporal structure, as demonstrated in the NARMA tasks,  
adjusting the value of $Q$ does not enhance the performance. 
However, in those cases, we found that increasing the value of $P$ can reduce the size of the reservoir without impairing performance. 
We have demonstrated that the number of neurons in the reservoir can be reduced by up to one tenth in the NMARMA10 task. 

Here, we briefly note the relationship between our work and the most relevant previous work \cite{Marquez2019Takens}.
In \cite{Marquez2019Takens}, the authors proposed a method that is similar to delay-state concatenation.
Their proposed model corresponds to the case when the number of additional connected past states is one (i.e., $P=1$). 
They observed that performance enhancement depends on the value of $Q$, as we showed in this paper. 
However, to the best of our knowledge, the dependence of the performance on the value of $P$ has not been reported.  
In addition, the other two proposed methods, drift-state concatenation and delay-state concatenation with transient states, 
are introduced for the first time in this paper. 
Moreover, the authors of \cite{Marquez2019Takens} explained the mechanism of their proposed method in terms of the delayed embedding theorem \cite{Takens1981detecting}. 
In contrast, we have provided a more intuitive explanation based on the IPC \cite{Dambre2012information}.  

Because the proposed methods do not assume a specific topology for the reservoir, 
they can readily be implemented in FPGAs and physical reservoir systems, such as photonic reservoirs \cite{Tanaka2019recent}. 
Therefore, the proposed methods could be an important set of techniques that facilitates the introduction of RC in edge computing. 

\ifnarxiv
\matmethods{

\subsection*{Training output weights}
The training procedures are the same as those for standard RC models \cite{Lukosevicius2012practical}.
The output weights are trained by minimizing 
\begin{flalign}
 \sum _{t=1}^{T} ||\bm{y}(t)- \bm{y}^\text{tc}(t)||_2 ^2. \label{eq:optimize}
\end{flalign}
Adding a regularization term $\| \hat{W}^\text{out}\|^2_2$ did not improve the performance in our case. 

\subsection*{Delay-state concatenation with transient states}

We inserted transient states in the RC system with delay-state concatenation as follows:
\begin{flalign}
\bm{x}(t) &= \tanh \Bigg(  W^\text{res}\bm{x}(t-1) \nonumber \\
&~~~~~~~~~~~~~+ W^\text{in}\bm{u}\left(\left\lfloor \frac{t}{N^\text{tran}+1} \right\rfloor \right)  \Bigg), \\
\bm{\hat{x}}(t) &= \left( \begin{array}{c}
\bm{x}(t) \\
\bm{x}(t-Q)\\
\vdots \\
\bm{x}(t-QP)
\end{array}
\right), \\
\bm{y}(t)& =W^\text{out}\bm{\hat{x}}\left((N^\text{tran}+1)t+N^\text{tran} \right), 
\end{flalign}
where $N^\text{tran}$ is the number of inserted transient states.

\subsection*{IPC}

Following the same procedure given in \cite{Dambre2012information}, 
the IPCs are calculated as follows: 
The total IPC is defined as 
\begin{flalign}
C^\text{total}&= \sum _{\{d_i \}} C\left(\{ d_i \}\right),
\end{flalign}
where $C^\text{total}$ is the total IPC and $C(\{ d_i \})$ is the IPC for a basis represented with a list $\{d_i\}=\{d_0, d_1, \dots \}$.
The list represents an orthogonal basis in Hilbert space.
We employed the following Legendre polynomials as the orthogonal basis: 
\begin{flalign}
y_{\{d_i\}}(t) = \prod _{i=0} ^{\tau_\text{max}} P_{d_i}\left( u(t-i)\right),
\end{flalign}
where $P_{d_i}(\cdot)$ is the $d_i$th-order Legendre polynomial and $u(t)$ is drawn from uniform distribution on $[-1,1]$.
A constant $\tau _\text{max}$ is the maximum delay, which must be large enough to converge the calculation.
In our simulations, we set $\tau _\text{max}$ to 50 for $\rho ^\text{in} = 0.1$ and to 25, otherwise.
Then, the IPC $C(\{d_i \})$ can be calculated as 
\begin{flalign}
C(\{d_i \}) = 1 - \frac{\left\langle |y(t)-y_{\{d_i\}}(t)|^2 \right\rangle}{\left\langle|y(t)-\langle y(t) \rangle|^2 \right\rangle},
\end{flalign}
where $\left\langle x(t)  \right\rangle := \frac{1}{T} \sum _{t=1}^{T} x(t)$.  
We set the simulation steps $T$ to $10^6$ in all experiments.
To avoid overestimation, the value of $C(\{d_i \})$ was set to zero when the value is less than threshold of $7N^\text{res}(P+1)\times 10^{-5}$.

We define the $k$th-order IPCs as
\begin{flalign}
C^{k \text{th}}&= \sum _{ \{d_i\} \in \Gamma ^k} C( \{ d_i \}), \\
\Gamma ^k &= \left\{\left\{d_i\right\} \left| \sum _{i=0}^{\tau _\text{max}} d_i = k \right .\right\}.
\end{flalign}
The $k$th-order IPC was decomposed into components corresponding to a subset of a basis whose maximum delay is $\tau$ as follows: 
\begin{flalign}
C^{k \text{th}} &= \sum _{\tau = 0}^{\tau _\text{max}}  C^{k \text{th}}_\tau, \\
C^{k \text{th}}_\tau &= \sum _{ \{d_i\} \in \Gamma ^k_\tau} C(X, \{ d_i \}), \\
\Gamma ^k_\tau &= \left\{\left\{d_i\right\} \left| \sum _{i=0}^{\tau _\text{max}} d_i = k, \max \{i| d_i \ge 1 \} = \tau \right. \right\}.
\end{flalign}

\subsection*{Dataset}

The $m$th-order generalized H\'{e}non-map \cite{Richter2002generalized} is given by 
\begin{flalign}
y^\text{tc}(t) = 1.76 - y^\text{tc}(t - m + 1)^2 - 0.1 y^\text{tc}(t -m) + \sigma(t) ~ (m\ge 2), \label{eq:Henon}
\end{flalign}
where $\sigma$ is Gaussian noise with zero mean and standard deviation of 0.05.
The inputs and outputs of the RC are the time series of an $n$-dimensional generalized H\'{e}non-map. 
The task is to predict one step forward $y(t+1)$ with past inputs $y(t), y(t-1), \dots$.
The NARMA time series is obtained with a nonlinear auto-regressive moving average as follows:
\begin{flalign}
y^\text{tc}(t) =& ~0.3y^\text{tc}(t-1) + 0.05y^\text{tc}(t-1) \sum _{i=1} ^{m} y^\text{tc}(t-i) \nonumber \\
&+ 1.5s(t-9) s(t) + 0.1, \label{eq:NARMA}
\end{flalign}
where $s(t)$ is drawn from uniform distribution of $[0, 0.5]$. 
The NARMA5 and NARMA10 time series correspond to the case when $m=5$ and $m=10$, respectively. 
The inputs of the RC are $s(t)$. The task is to predict $y(t)$ from the inputs $s(t)$.

For both tasks, we used 2,000 steps as a training dataset 
and used 3,000 steps as a test dataset. 
We removed the first 200 steps (free run) both during the training and test phases to avoid the effects of the initial conditions in the reservoirs \cite{Lukosevicius2012practical}. 
We evaluated the performance based on the normalized mean-squared error (NMSE) during the test phase following: 
\begin{flalign}
\text{NMSE} = \frac{\left\langle |y(t)-y^\text{tc}(t)|^2 \right\rangle}{\left\langle|y(t)-\langle y(t) \rangle|^2 \right\rangle},
\end{flalign}
where $\left\langle x(t)  \right\rangle = \frac{1}{T} \sum _{t=1}^{T} x(t)$. 
We averaged the NMSEs over 10 trials.
For each iteration, the dataset and connection matrix of the reservoir were generated using their corresponding probabilistic distributions.

}

\showmatmethods{} 

\else
\section*{Materials and Methods}
\subsection*{Training output weights}
The training procedures are the same as those for standard RC models \cite{Lukosevicius2012practical}.
The output weights are trained by minimizing 
\begin{flalign}
 \sum _{t=1}^{T} ||\bm{y}(t)- \bm{y}^\text{tc}(t)||_2 ^2. \label{eq:optimize}
\end{flalign}
Adding a regularization term $\| \hat{W}^\text{out}\|^2_2$ did not improve the performance in our case. 

\subsection*{Delay-state concatenation with transient states}

We inserted transient states in the RC system with delay-state concatenation as follows:
\begin{flalign}
\bm{x}(t) &= \tanh \Bigg(  W^\text{res}\bm{x}(t-1) \nonumber \\
&~~~~~~~~~~~~~+ W^\text{in}\bm{u}\left(\left\lfloor \frac{t}{N^\text{tran}+1} \right\rfloor \right)  \Bigg), \\
\bm{\hat{x}}(t) &= \left( \begin{array}{c}
\bm{x}(t) \\
\bm{x}(t-Q)\\
\vdots \\
\bm{x}(t-QP)
\end{array}
\right), \\
\bm{y}(t)& =W^\text{out}\bm{\hat{x}}\left((N^\text{tran}+1)t+N^\text{tran} \right), 
\end{flalign}
where $N^\text{tran}$ is the number of inserted transient states.

\subsection*{IPC}

Following the same procedure given in \cite{Dambre2012information}, 
the IPCs are calculated as follows: 
The total IPC is defined as 
\begin{flalign}
C^\text{total}&= \sum _{\{d_i \}} C\left(\{ d_i \}\right),
\end{flalign}
where $C^\text{total}$ is the total IPC and $C(\{ d_i \})$ is the IPC for a basis represented with a list $\{d_i\}=\{d_0, d_1, \dots \}$.
The list represents an orthogonal basis in Hilbert space.
We employed the following Legendre polynomials as the orthogonal basis: 
\begin{flalign}
y_{\{d_i\}}(t) = \prod _{i=0} ^{\tau_\text{max}} P_{d_i}\left( u(t-i)\right),
\end{flalign}
where $P_{d_i}(\cdot)$ is the $d_i$th-order Legendre polynomial and $u(t)$ is drawn from uniform distribution on $[-1,1]$.
A constant $\tau _\text{max}$ is the maximum delay, which must be large enough to converge the calculation.
In our simulations, we set $\tau _\text{max}$ to 50 for $\rho ^\text{in} = 0.1$ and to 25, otherwise.
Then, the IPC $C(\{d_i \})$ can be calculated as 
\begin{flalign}
C(\{d_i \}) = 1 - \frac{\left\langle |y(t)-y_{\{d_i\}}(t)|^2 \right\rangle}{\left\langle|y(t)-\langle y(t) \rangle|^2 \right\rangle},
\end{flalign}
where $\left\langle x(t)  \right\rangle := \frac{1}{T} \sum _{t=1}^{T} x(t)$.  
We set the simulation steps $T$ to $10^6$ in all experiments.
To avoid overestimation, the value of $C(\{d_i \})$ was set to zero when the value is less than threshold of $7N^\text{res}(P+1)\times 10^{-5}$.

We define the $k$th-order IPCs as
\begin{flalign}
C^{k \text{th}}&= \sum _{ \{d_i\} \in \Gamma ^k} C( \{ d_i \}), \\
\Gamma ^k &= \left\{\left\{d_i\right\} \left| \sum _{i=0}^{\tau _\text{max}} d_i = k \right .\right\}.
\end{flalign}
The $k$th-order IPC was decomposed into components corresponding to a subset of a basis whose maximum delay is $\tau$ as follows: 
\begin{flalign}
C^{k \text{th}} &= \sum _{\tau = 0}^{\tau _\text{max}}  C^{k \text{th}}_\tau, \\
C^{k \text{th}}_\tau &= \sum _{ \{d_i\} \in \Gamma ^k_\tau} C(X, \{ d_i \}), \\
\Gamma ^k_\tau &= \left\{\left\{d_i\right\} \left| \sum _{i=0}^{\tau _\text{max}} d_i = k, \max \{i| d_i \ge 1 \} = \tau \right. \right\}.
\end{flalign}

\subsection*{Dataset}

The $m$th-order generalized H\'{e}non-map \cite{Richter2002generalized} is given by 
\begin{flalign}
y^\text{tc}(t) = 1.76 - y^\text{tc}(t - m + 1)^2 - 0.1 y^\text{tc}(t -m) + \sigma(t) ~ (m\ge 2), \label{eq:Henon}
\end{flalign}
where $\sigma$ is Gaussian noise with zero mean and standard deviation of 0.05.
The inputs and outputs of the RC are the time series of an $n$-dimensional generalized H\'{e}non-map. 
The task is to predict one step forward $y(t+1)$ with past inputs $y(t), y(t-1), \dots$.
The NARMA time series is obtained with a nonlinear auto-regressive moving average as follows:
\begin{flalign}
y^\text{tc}(t) =& ~0.3y^\text{tc}(t-1) + 0.05y^\text{tc}(t-1) \sum _{i=1} ^{m} y^\text{tc}(t-i) \nonumber \\
&+ 1.5s(t-9) s(t) + 0.1, \label{eq:NARMA}
\end{flalign}
where $s(t)$ is drawn from uniform distribution of $[0, 0.5]$. 
The NARMA5 and NARMA10 time series correspond to the case when $m=5$ and $m=10$, respectively. 
The inputs of the RC are $s(t)$. The task is to predict $y(t)$ from the inputs $s(t)$.

For both tasks, we used 2,000 steps as a training dataset 
and used 3,000 steps as a test dataset. 
We removed the first 200 steps (free run) both during the training and test phases to avoid the effects of the initial conditions in the reservoirs \cite{Lukosevicius2012practical}. 
We evaluated the performance based on the normalized mean-squared error (NMSE) during the test phase following: 
\begin{flalign}
\text{NMSE} = \frac{\left\langle |y(t)-y^\text{tc}(t)|^2 \right\rangle}{\left\langle|y(t)-\langle y(t) \rangle|^2 \right\rangle},
\end{flalign}
where $\left\langle x(t)  \right\rangle = \frac{1}{T} \sum _{t=1}^{T} x(t)$. 
We averaged the NMSEs over 10 trials.
For each iteration, the dataset and connection matrix of the reservoir were generated using their corresponding probabilistic distributions.

\fi

\ifnarxiv
\acknow{
The authors would like to thank Makoto Ikeda, Hiromitsu Awano, and Gouhei Tanaka for the fruitful discussion.  
This work was partially supported by the ``Brain-Morphic AI to Resolve Social Issues'' project at UTokyo, the NEC Corporation, and 
AMED (JP20dm0307009).
}
\showacknow{} 
\else

\section*{Acknowledgment}
The authors would like to thank Makoto Ikeda, Hiromitsu Awano, and Gouhei Tanaka for the fruitful discussion.  
This work was partially supported by the ``Brain-Morphic AI to Resolve Social Issues'' project at UTokyo, the NEC Corporation, and 
AMED (JP20dm0307009).
\fi

\ifnarxiv
\else
\bibliographystyle{junsrt}
\fi

\bibliography{myBib}

\end{document}